\documentclass[runningheads]{llncs}
\usepackage[T1]{fontenc}
\usepackage[dvipsnames,table]{xcolor} 
\usepackage{xspace}
\usepackage{graphicx}
\usepackage{amssymb}
\usepackage{amsmath}
\usepackage{ dsfont }
\usepackage{multirow}
\usepackage{lmodern}
\definecolor{lightgray}{gray}{0.95}
\definecolor{midgray}{gray}{0.55}
\definecolor{steelblue}{HTML}{4D82B7}
\usepackage{makecell}
\usepackage{nicefrac}
\usepackage[export]{adjustbox}
\usepackage{mathtools}
\usepackage{booktabs}
\newcommand{\methodname}{Low-rank Adaptation for VQ-VAE\xspace}
\newcommand{\methnam}{LRVQ\xspace}

\DeclarePairedDelimiterX{\infdivx}[2]{(}{)}{%
  #1\;\delimsize\|\;#2%
}

\newcommand{\kld}[2]{\ensuremath{D_{KL}\infdivx{#1}{#2}}\xspace}

\newcommand{\ie}{\textit{i.e.},~}
\newcommand{\eg}{\textit{e.g.},~}
\newcommand{\tit}[1]{\smallbreak\noindent\textbf{#1 }}

\newcommand{\tinysum}{{\scriptsize  \sum}}
\newcommand{\ctx}{\operatorname{ctx}}
\newcommand{\enc}{\operatorname{E}}
\newcommand{\dec}{\operatorname{G}}
\usepackage{graphicx}
\usepackage{amssymb}
\usepackage{amsfonts}
\usepackage{amsmath}
\usepackage{color}
\begin{document}
\title{Trajectory Forecasting through Low-Rank Adaptation of Discrete Latent Codes}
\titlerunning{Low-Rank Adaptation of Discrete Latent Codes}
\author{Riccardo Benaglia\inst{1,2}\orcidID{0009-0002-8788-1156} \and
Angelo Porrello\inst{1}\orcidID{0000-0002-9022-8484} \and
Pietro Buzzega\inst{1}\orcidID{0000-0002-6516-6373} \and Simone Calderara\inst{1}\orcidID{0000-0001-9056-1538} \and
Rita Cucchiara\inst{1}\orcidID{0000-0002-2239-283X}}
\authorrunning{R. Benaglia et al.}
\institute{AImageLab - University of Modena and Reggio Emilia, Modena, Italy\\
\email{name.surname@unimore.it} \and
Ammagamma S.r.l.\\
}
\maketitle
\begin{abstract}
Trajectory forecasting is crucial for video surveillance analytics, as it enables the anticipation of future movements for a set of agents, \eg basketball players engaged in intricate interactions with long-term intentions. Deep generative models offer a natural learning approach for trajectory forecasting, yet they encounter difficulties in achieving an optimal balance between sampling fidelity and diversity. We address this challenge by leveraging Vector Quantized Variational Autoencoders (VQ-VAEs), which utilize a discrete latent space to tackle the issue of posterior collapse. Specifically, we introduce an instance-based codebook that allows tailored latent representations for each example. In a nutshell, the rows of the codebook are dynamically adjusted to reflect contextual information (\ie past motion patterns extracted from the observed trajectories). In this way, the discretization process gains flexibility, leading to improved reconstructions. Notably, instance-level dynamics are injected into the codebook through low-rank updates, which restrict the customization of the codebook to a lower dimension space. The resulting discrete space serves as the basis of the subsequent step, which regards the training of a diffusion-based predictive model. We show that such a two-fold framework, augmented with instance-level discretization, leads to accurate and diverse forecasts, yielding state-of-the-art performance on three established benchmarks.
\keywords{Trajectory forecasting \and Vector Quantization.}
\end{abstract}
\section{Introduction}
Trajectory forecasting finds applications in video surveillance~\cite{li2021temporal}, multi-object tracking~\cite{mancusi2023trackflow,dendorfer2022quo}, behavioural analysis~\cite{rudenko2020human}, and intrusion detection~\cite{sun2020pidnet}. The goal is to predict the future paths of a set of agents from a few observations of their motion. The prediction can incorporate the interactions between pedestrians~\cite{huang2019stgat,sun2019stochastic,monti2022many}, or visual attributes of the environment they move within~\cite{dendorfer2021mg}. 

As multiple plausible paths can be forecast, trajectory prediction reveals an uncertain and multi-modal nature. To achieve this, recent data-driven approaches~\cite{mangalam2020not,gupta2018social,ivanovic2019trajectron} lean toward a stochastic formulation that places a distribution over the future trajectory, rather than a single estimated path with $100\%$ certainty (\ie \textit{deterministic} approaches~\cite{monti2022many}). In doing so, recent stochastic methods take advantage of the latest breakthroughs in deep generative modeling for image generation. For example,~\cite{gupta2018social,sadeghian2019sophie} resorted to Generative Adversarial Networks, while~\cite{yuan2019diverse,ivanovic2019trajectron,tang2019multiple} borrowed ideas from the class of variational methods.

One of the hindrances toward the application of variational approaches is the \textit{posterior collapse} issue: \ie when the latent variables collapse to the prior becoming uninformative; as a consequence, the decoder learns to ignore them. This translates into a model with undermined generative capabilities, wherein its predictions are distributed on a single path (\eg, the most trivial one) with low uncertainty. A similar tendency (\textit{mode collapse}) has been observed in adversarial networks, and has been addressed through burdensome learning objectives promoting variety~\cite{gupta2018social,sadeghian2019sophie}, or by devising multiple generator networks~\cite{dendorfer2021mg}.

In the field of image generation, \textbf{Vector Quantized Variational Autoencoders~\cite{van2017neural}} (VQ-VAEs) have proven to mitigate posterior collapse. VQ-VAEs models avoid the hand-crafted Gaussian prior distribution; differently, they build upon a learnable categorical prior, thereby yielding a discrete latent space. The symbols of this space are the keys of a fixed-size dictionary (\textbf{codebook}), whose values are learnable latent codes. Thanks to the resulting increased flexibility, VQ-VAEs embody a promising paradigm for trajectory forecasting.

In this respect, our main contribution regards the content of the VQ-VAE codebook. In particular, while the original formulation devises a single codebook shared across all examples, we propose to dynamically adjust its values based on the \textit{context} of each example, leading to an \textbf{instance-based} codebook. We refer as \textit{context} to the set of historical information related to each agent, namely the past steps of its trajectory as well as its interactions with nearby agents. In this way, we aim to encourage even more flexibility during the discretization process, as distinct motion patterns can be discretized with varying granularity. 

Moreover, we envision the customization of the codebook as an \textbf{adaptation} of the shared original VQ-VAE codebook. By doing so, our goal is to strike a balance between per-instance customization and the emergence of cross-instance concepts that are relevant across multiple examples. In practice, we draw inspiration from recent advances in Parameter Efficient Fine Tuning and represent the dynamic adjustments to the codebook as \textbf{low-rank updates} of its values (see Fig.~\ref{fig:vqvae}). We show that such a modeling constraint improves the representation capabilities of the learned latent space, thereby encoding additional information and facilitating the reconstruction task. The traditional subsequent stage in VQ-VAEs involves fitting the distribution on the discrete latent codes. In this respect, we make use of a vector-quantized diffusion model~\cite{gu2022vector} to learn the implicit prior, departing from existing approaches~\cite{van2017neural,esser2021taming} that rely on autoregressive priors, which are more susceptible to issues related to error accumulation.

The contributions are \textit{i)} to the best of our knowledge, we are the first leveraging VQ-VAEs in a trajectory generation task; \textit{ii)} we introduce a novel instance-based codebook based on low-rank modeling; \textit{iii)} we achieve SOTA performance on three established benchmarks (Stanford Drone~\cite{robicquet2016learning}, NBA~\cite{NBA} and NFL~\cite{NFL}).
\section{Related Work}
\label{sec:related:traj} 
The traditional approach to trajectory prediction considers solely the past movements of the agent~\cite{becker2018red}. However, its motion is likely to be influenced by the motions of other agents (\eg to avoid collisions or to perform coordinate actions). The first approaches took into account social behaviors through hand-crafted relations, energy-based features, or rule-based models~\cite{antonini2006discrete,pellegrini2009you}. In recent years, the focus has shifted towards data-driven approaches~\cite{alahi2016social,gupta2018social}, leveraging deep models to extract social information~\cite{huang2019stgat,sun2019stochastic}. Others, instead, rely on the attention mechanism, which has proven highly effective at capturing interactions within tokenized data~\cite{monti2022many}. For example, \cite{huang2019stgat} employs a graph-based attention mechanism to model human interactions, while~\cite{monti2022many} utilizes a social-temporal attention module to capture temporal relationships between consecutive time steps and interpersonal interactions occurring among agents.

Given the inherent uncertainty and multi-modal characteristics of future trajectories, recent approaches embrace a deep probabilistic framework to model their distribution. S-GAN~\cite{gupta2018social} leverages a conditional Generative Adversarial Network (GAN)~\cite{goodfellow2014generative}, while the authors of SoPhie~\cite{sadeghian2019sophie} extend GANs to incorporates visual and social interaction components. Other works utilize conditional Variational Autoencoders (VAE)~\cite{kingma2013auto} for multimodal pedestrian trajectory prediction, including~\cite{yuan2019diverse,ivanovic2019trajectron,tang2019multiple,salzmann2020trajectron++,yuan2021agentformer}. Trajectron\texttt{++}~\cite{salzmann2020trajectron++} employs a VAE and represents agents' trajectories in a graph-structured recurrent neural network, while PECNet~\cite{mangalam2020not} integrates VAEs and goal conditioning. However, both GAN and VAE-based methods grapple with collapsing issues in trajectory generation, necessitating burdensome countermeasures~\cite{thiede2019analyzing}. Ultimately, the work by~\cite{gu2022stochastic} pioneers the utilization of denoising diffusion models~\cite{ho2020denoising} within the trajectory prediction framework, marking a significant advancement in this domain.
\subsubsection{Vector Quantization Models.}
\label{sec:related:vqvae} 
Vector Quantized Variational Autoencoders~\cite{van2017neural} address posterior collapse by replacing the continuous latent space of VAEs with a discrete set of codewords. Starting from pioneering works, which showed the potential of these models in image generation~\cite{van2017neural,razavi2019generating}, recent studies focused on improving the two fundamental stages: the codebook learning and the discrete prior learning~\cite{kolesnikov2022uvim}. In this respect, SQ-VAE~\cite{takida2022sq} replaces deterministic quantization with a pair of stochastic dequantization and quantization processes. To create a more comprehensive codebook,~\cite{esser2021taming} supplements the original training losses of VQ-VAE with adversarial training. Additionally,~\cite{zhang2023regularized} adopts a masking strategy during training and introduces prior distribution regularization to mitigate issues related to low-codebook utilization.

The advances regarding discrete prior learning involve architectural modifications~\cite{esser2021taming} and a critical reevaluation of autoregression. \cite{tang2022improved} employs a discrete diffusion architecture to model code prediction, while MaskGIT~\cite{chang2022maskgit} utilizes a bidirectional transformer decoder. This decoder generates all tokens of an image simultaneously and iteratively refines the image based on the preceding generation. In this paper, we condition the codebook on historical instance-level information while preserving the discrete nature of the latent space.
\section{Preliminaries}
\label{sec:background}
We denote the future trajectory as $y \in \mathbb{R}^{T \times d}$, where $T$ is the number of future time steps and $d$ is the input channel dimension. When dealing with pedestrians, their trajectories are projected into the 2D bird's-eye view (so $d=2$). The predicted trajectory $\hat{y}$ is generated by a learnable model, fed with a set of conditioning information:~\textit{i)} the observed trajectory $x \in \mathbb{R}^{T_{p} \times d}$ of the agent, \ie the coordinates observed at previous $T_{p}$ steps, and~\textit{ii)} a set of neighboring trajectories denoted as $\mathcal{X} = \{x_{1}, x_{2}, \dots , x_{N}\}$. We define neighbors of an agent as all agents within the same scene, without imposing any distance threshold.

\smallskip
\tit{Vector Quantization.} Standard VAEs~\cite{kingma2013auto} employ \textit{i)} an \textbf{encoder} $\enc \equiv \enc(y | \theta_{\enc})$ that, given input $y$, outputs a parametric posterior distribution $q(z|y)$ over latent variable $z$; \textit{ii)} a \textbf{decoder} $\dec \equiv \dec(z | \theta_{\dec})$ that provides the reconstruction of the input data as $p_{\theta_{\dec}}(y|z)$. The posterior $q(z|y)$ is encouraged to conform to a standard Gaussian prior distribution $p(z)$, which could lead to over-regularized representations (posterior collapse). VQ-VAEs~\cite{van2017neural} extend VAEs by employing discrete latent variables and Vector Quantization (VQ)~\cite{gray1998quantization}. In particular, both posterior and prior distributions are categorical, and their samples provide indices for a learned \textbf{embedding table} $e \in \mathbb{R}^{C \times D}$, which consists of $C$ static $D$-dimensional latent vectors. As outlined in the following paragraphs, the training of VQ-VAEs is divided into \textit{learning the codebook} and \textit{fitting the categorical prior}.

\smallskip
\tit{First Stage.} Given the input $y \in \mathbb{R}^{T \times d}$, the encoder provides a continuous representation $z \in \mathbb{R}^{T\times D}$, where $z_t \in \mathbb{R}^{D} \text{ with } t \in \{1, 2, \dots, T\}$ and $D$ indicates the dimension of the latent space. Then, the VQ-VAE characterizes the posterior as a joint distribution over $T$ independent \textbf{categorical} variables $q(c_1, c_2, \dots, c_T|y)$ (one for each latent). Each marginal $q(c_t|y)$ is determined by matching each element of the encoding sequence $z_t$ with the \textbf{nearest} vector in the codebook $e$:
\begin{equation}
q(c_t|y) = \underbrace{\mathcal{C}(p_1, p_2, \dots, p_{C})}_{[0, \dots, 0, 1, 0, \dots, 0]} \text{ s.t. } p_c = \begin{cases} 1 & \text{if } c = \operatorname{argmin}_{c' \in \{1, 2, \dots, C\}} \|z_t - e_{c'}\|_2^2 \\ 0 & \text{otherwise.} \end{cases}
\end{equation}
Notably, the posterior distribution is \textit{deterministic} and not stochastic as for VAEs: hence, we can \textit{draw} a sample $z^q \equiv z^q(y)$ from the posterior distribution by \textbf{selecting} the corresponding rows of the codebook, as follows:
\begin{equation}
\label{eq:nn_match}
\begin{aligned}
z^q &= [e_{c_1}, e_{c_2}, \dots, e_{c_T}]\\
c_t \sim q(c_t|y) & \implies c_t = \operatorname{argmax} q(c_t|y).
\end{aligned}
\end{equation}
The subsequent step regards the decoder $\dec$, which reconstructs $\Hat{y}$ from the sampled latent vector. During training, the first stage optimizes the following loss:
\begin{equation}
\mathcal{L}_{\text{FS}} = \underbrace{\log p_{\theta_{\dec}}(y|z^q)}_{\text{rec. error \eg MSE}}+ \tinysum_{t} \underbrace{\|\texttt{sg}[z_t] - e_{c_t}\|^2}_{\text{embedding loss}} + \tinysum_{t} \underbrace{\|z_t - \texttt{sg}[e_{c_t}])\|^2}_{\text{commitment loss}}, \label{eq:first_stage}
\end{equation}
where $\texttt{sg}$ is a shortcut for the $\texttt{stopgradient}$ operator, which stops backpropagation from that computational node backward. The second term encourages the quantized latent vectors to be as close as possible to the nearest codeword, while the third one encourages the encoder to be \textit{committed} to the chosen codeword.

\tit{Second Stage.}The goal here is to learn a parametric model $p_{\theta_p}(c_1, c_2, \dots, c_T)$ -- termed \textit{categorical prior} -- which allows to draw new samples from the latent space. During this phase, the modules of the VQ-VAE are no longer subject to learning. Given the trained encoder, each training example $y$ is embedded into a sequence of indices, built by relating each latent vector to the nearest row of the codebook (as in Eq.~\ref{eq:nn_match}). On top of that, the generative model targets the generating process $p(c_1, c_2, \dots, c_T)$ of the discrete latent codes, and optimizes the following Maximum Likelihood Estimation (MLE) training objective: 
\begin{equation}
    \mathcal{L}_{\text{SS}} = \mathbb{E}_{\substack{c_1, \dots, c_T \\ c_t \sim q(c_t|y)}} [-\log p_{\theta_p} (c_1, c_2, \dots, c_T)]. \label{eq:second_stage}
\end{equation}
\section{\methodname}
\label{sec:vq_traj}
\begin{figure}[t]
\centering
\includegraphics[width=0.99\textwidth]{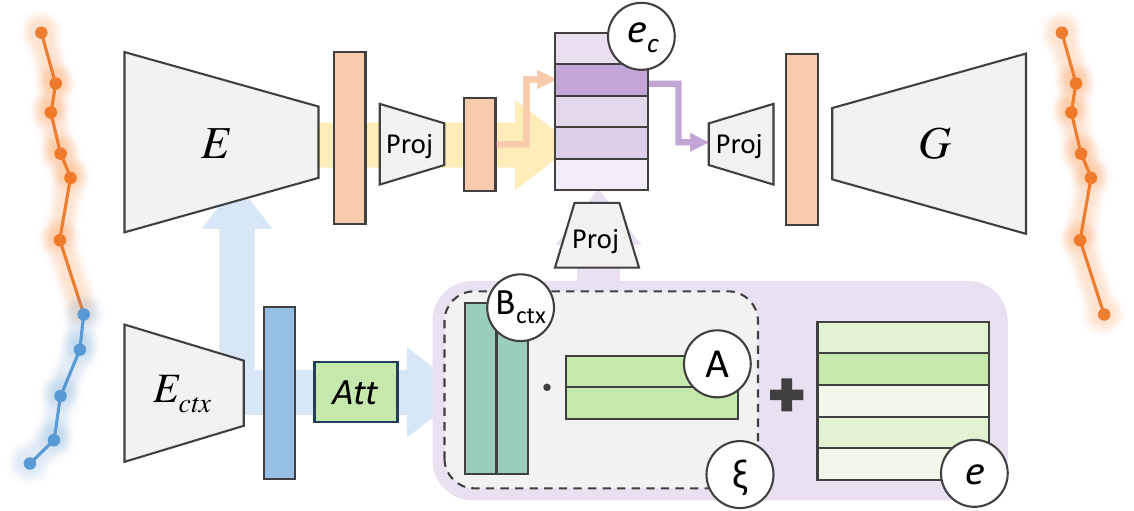}
    \caption{Overview of our approach to trajectory prediction, based on Vector Quantization and Low-Rank adaptation of the codebook (highlighted in the purple box).}
    \label{fig:vqvae}
\end{figure}
We herein present our approach to trajectory prediction, which we name \methnam, depicted in Fig.~\ref{fig:vqvae}. Briefly, we exploit VQ-VAEs to encode the future trajectory $y$ of a given agent. On top of that, the following main novelties are introduced:
\begin{itemize}
    \item We extend VQ-VAE to predict a trajectory coherent with the observed historical trend. To do so, we feed \textbf{additional contextual} information to the VQ-VAE, conditioning both the prior and the posterior distributions. The contextual information consists of the past observed trajectory $x$, and a summary of the interactions between the agent and its neighbours. The structure of the resulting quantization model is presented in Sec.~\ref{sec:vqvae}.
    \item To encourage further \textbf{flexibility}, the codebook itself is conditioned on the additional contextual information (see Sec.~\ref{sec:codebook}). As discussed later, the context is introduced by devising a \textbf{low-rank} adjustment to the codebook.
    \item To avoid the error accumulation and the \textit{unidirectional bias} problem, typical of auto-regressive methods~\cite{gu2022vector}, we make use of a \textbf{discrete diffusion model} for the generation of the sequence of indices (see Sec.~\ref{sec:categorical_prior}). We also introduce a \textbf{new sampling technique}, based on the k-means clustering algorithm, to produce better and more consistent generations (see Sec.~\ref{sec:multimodal}).
\end{itemize}
\vspace{-0.4em}
\subsection{Trajectory Forecasting with VQ-VAEs}
\label{sec:vqvae}
Formally, our VQ-VAE can be summarized as:
\begin{subequations}
\begin{align}
h_{\ctx} &= \enc_{\ctx}([x, \mathcal{X}]) &\ \text{(context encoding)} \label{eq:encoding} \\
z^q &= \enc(y, [\mathcal{Y}, h_{\ctx}]) \label{eq:encoding_future} &\ \text{(encoding)} \\
\hat{y} &= \dec(z^q, \mathcal{Z}^q), &\ \text{(decoding)}\label{eq:decode}
\end{align}
\end{subequations}
where $\mathcal{X}$, $\mathcal{Y}$ and $\mathcal{Z}^q$ represent respectively the past, the future, and the latent quantized representation of the nearby agents' trajectories (see Sec.~\ref{sec:background}). The modules $\enc_{\ctx}(\cdot)$, $\enc(\cdot)$, $\dec(\cdot)$ are three neural networks, each of which exploits social-temporal transformer~\cite{monti2022many} to account for social-temporal relations.

In particular, a contextual encoder $\enc_{\ctx}(\cdot)$ computes hidden features $h_{\ctx} \in \mathbb{R}^{Tp \times D}$ that summarize both the past trend $x \in \mathbb{R}^{T_p \times 2}$ of the trajectory and spatial interactions (Eq.~\ref{eq:encoding}). The function $\enc(\cdot)$ plays the role of the VQ-VAE encoder, transforming the future trajectory $y$ into a discrete representation $z^q \in \mathbb{R}^{T \times D}$ (see Eq.~\ref{eq:encoding_future}). To condition the model on historical information, the encoder is fed also with the hidden contextual information $h_{\ctx}$; in detail, a tailored cross-attention layer is devised to mix future and past information. Finally, in step~(\ref{eq:decode}) we achieve the estimated future trajectory $\hat{y} \in \mathbb{R}^{T \times 2}$ through the decoder $\dec(\cdot)$.

As well as traditional VQ-VAEs, we employ Mean Squared Error (MSE) as our reconstruction term between the ground truth and predicted trajectory.
\vspace{-0.4em}
\subsection{Instance-based Codebook}
\label{sec:codebook}
The codebook plays a crucial role in VQ-VAEs and can cause instabilities during optimization. For instance, the uneven utilization of the vectors of the codebook is a factor that may lead to inefficiencies in representation learning. This imbalance often results in certain elements of the codebook being underutilized, while others never match with real-valued embeddings. To mitigate these issues, the authors of~\cite{yu2021vector} resort to reducing the latent-space dimensionality, showing that it leads to a condensed but richer codebook. In practice, before quantization, each vector $z$ is projected from $\mathbb{R}^{D}$ to a lower-dimension space $D_r \ll D$. In the following, we will refer to this strategy as \textbf{static codebook}, to distinguish it from our proposal that instead leverages dynamic cues.

Our idea is to modify the content of the codebook, such that it reflects the motion observed in the past trajectory. The intuition is that different motion styles (\eg straight \textit{vs.} curvilinear) could prefer distinct latent codes and discretization strategies. On this basis, we exploit again the contextual features $h_{\ctx}$ to generate an \textbf{instance-based codebook} $\xi = f_{\xi}(\cdot, h_{\ctx})$, computed through a tailored learnable module $f_{\xi}$. The latter shares the same design of the above-described encoding networks and hence builds upon social-temporal transformers~\cite{monti2022many}. Afterwards, we combine static and instance-based codebooks by means of summation, thus obtaining a \textbf{conditioned} codebook $e_c$:
\begin{equation}
\label{eq:final_condition_codebook}
    e_c = \texttt{l2\_norm}(e) + \lambda_\xi \texttt{l2\_norm}(\xi)
\end{equation}
where $\texttt{l2\_norm}$ indicates the row-wise l2-normalization $\nicefrac{v}{\lVert v \rVert_2}$ and $\lambda_\xi$ is an hyperparameter that weighs the sum. We leverage normalizing layers to ensure that the two components contribute almost equally to the final embedding table.

Moreover, the way we define the codebook draws inspiration from the successes of low-rank adaptation~\cite{hu2021lora} for fine-tuning Large Language Models (LLMs). Namely, we opt for a \textit{low-rank characterization} of $f_\xi$, which means that the instance-driven modifications to the static codebook lie on a lower-dimensional manifold of the parameter space. We hence define the instance-based codebook $\xi$ as a matrix product of two low-rank matrices $B_{\ctx}$ and $A$, as follow:
\begin{equation}
\begin{aligned}
B_{\ctx} &= f_{\xi}(B,h_{\ctx}) \quad \text{where}~B, B_{\ctx} \in \mathbb{R}^{D \times r}\\
\xi &= B_{\ctx} A \quad \text{where } A \in \mathbb{R}^{r \times C}. \\ 
\end{aligned}
\label{eq:codebook_sum}
\end{equation}
Considering $B$ as a set of learnable tokens, $f_\xi$ adopts cross attention between the conditioning information $h_{\ctx}$ and $B$ to create an instance-based $B_{\ctx}$.
\vspace{-0.4em}
\subsection{Diffusion-based Categorical Prior}
\label{sec:categorical_prior}
As previously mentioned, the second main stage regards the training of the parametric categorical prior $p_{\theta_p}(c|x, \mathcal{X})$ (note that the $p_{\theta_p}$ is also conditioned on historical information), where $c=\{c_1, c_2, \dots, c_T\}$. Notably, the learned prior serves to forecast the future trajectory $y$ at inference time, when the posterior distribution of $y$ is not available. Sec.~\ref{sec:multimodal} provides a detailed description of the sampling procedure, while the rest of this section describes the architectural and training aspects of the categorical prior. 

We borrow the design of the categorical prior from the framework of Denoising Diffusion Probabilistic Models (DDPMs). In particular, we employ vector-quantized diffusion models~\cite{gu2022vector}, as they naturally handle discrete distributions. Notably, the application of DDPMs allows one to learn the categorical prior without the need for autoregressive modeling, as commonly employed in many existing approaches~\cite{van2016pixel,esser2021taming}. In the context of trajectory prediction, we view the adoption of a non-autoregressive model as an additional strength. On the one hand, auto-regressive methods can leverage the inherent inductive bias of time-series data, where consecutive time steps relate to each other. However, this often results in error accumulation issues and in the so-called \textit{unidirectional bias}~\cite{gu2022vector}, which blurs contextual information that flows in a direction not coherent with the chosen auto-regressive order. In the task under consideration, this means that auto-regressive approaches may struggle to leverage cues emerging in later moments of the trajectory, as \textit{the goal} or the long-range intention of the agent. These crucial aspects of trajectory prediction~\cite{mangalam2020not} could be better addressed by the approach proposed in this work, which is \textbf{order-free} and capable of capturing multiple plausible trends.

Formally, we define $q^{\text{\textit{diff}}}$ as the diffusion process that injects incremental noise to the token sequence $c$ for $\Psi$ diffusion steps. Instead, $p_{\theta}^{\text{\textit{diff}}}$ is the denoising process that gradually reduces the noise of the noised sequence. The parameters $\theta$ of the denoising module are trained with the variational lower bound~\cite{sohl2015deep}:
\begin{subequations}
    \begin{align}
        &\mathcal{L}_{\mathrm{vlb}} = \mathcal{L}^0 + \mathcal{L}^1 + \cdots + \mathcal{L}^{\Psi-1} + \mathcal{L}^{\Psi}, 
        \\ 
        &\mathcal{L}^{\psi} = \kld{q^{\text{\textit{diff}}}{({c}^{\psi}|{c}^{\psi-1})}}{p_\theta^{\text{\textit{diff}}}({c}^{\psi}|{c}^{\psi+1},\widehat{\mathcal{C}}^{\psi}, x, \mathcal{X})}, 
        \\
        \label{eqn:aux_loss}
        &\mathcal{L}^{c^0} = -\log p_{\theta}^{\text{\textit{diff}}}({c}^0|{c}^\psi,\widehat{\mathcal{C}}^{\psi}, x, \mathcal{X}),
    \end{align}
\end{subequations}
where we use $x$, $\mathcal{X}$ and $\widehat{\mathcal{C}}^{\psi}$ -- the token sequence of neighboring agents at diffusion step $\psi$ -- as conditioning information during denoising. (\ref{eqn:aux_loss}) is an auxiliary objective encouraging the prediction of a noiseless token $s_0$. The loss function:
\begin{equation}
    \mathcal{L} \gets \begin{cases}
			\mathcal{L}^0, &\text{if } \psi = 1 \\
			\mathcal{L}^{\psi-1} + \lambda \mathcal{L}^{c^0} &\text{otherwise}.
		\end{cases}
\end{equation}
We refer to~\cite{gu2022vector} for more exhaustive details on the diffusion steps and the prior. 

\tit{Generation.} At inference time, the past and social information is encoded using $\mathrm{E_{ctx}}$ and then passed to the diffusion process $p_{\theta}^{\text{\textit{diff}}}$. The latter, after $\Psi$ denoising steps, provides a (denoised) sequence of $T$ indices $\hat{c} \in \mathbb{R}^{T}$. These indices represent the encoding of the future unobserved trajectory; therefore, we used them to select the proper elements of the codebook $e_c$, thus allowing us to create a quantized sequence representation $z^q$. Then $z^q$ undergoes decoding through the VQ-VAE decoder $\dec$, which finally yields the generation of trajectories $\hat{y}$.
\subsection{Enforcing Effective Multi-modal Forecasting}
\label{sec:multimodal}
\begin{figure}[t]
\centering
    \includegraphics[width=0.95\textwidth]{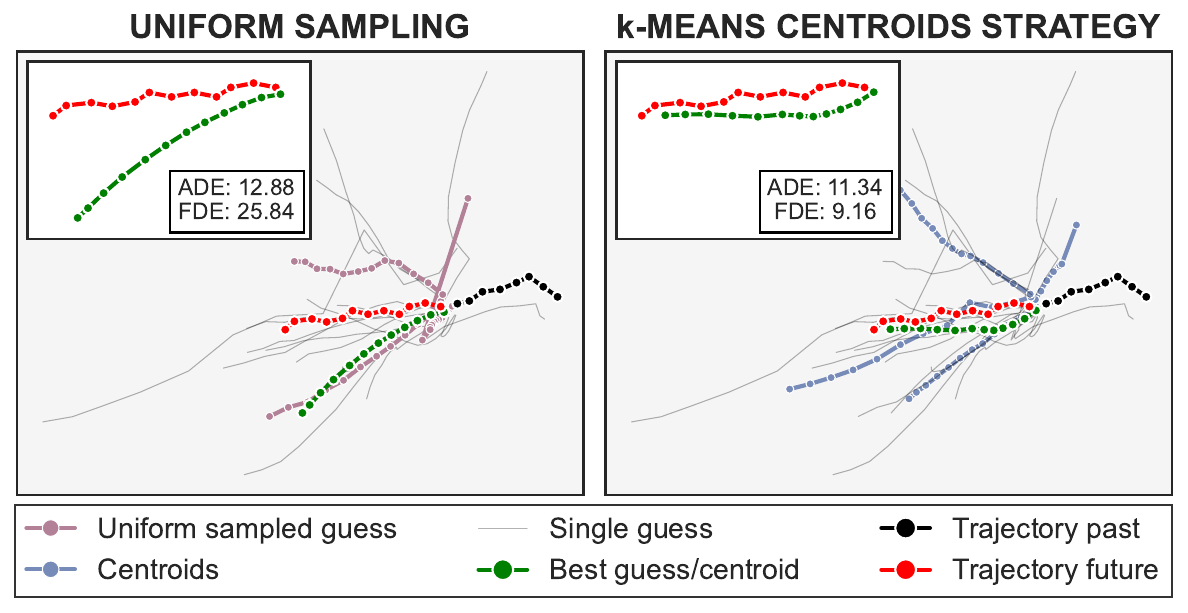}
    \caption{Comparison between the $K = 5$ samples obtained from a uniform sampling strategy (on the left) and the ones given as output from the proposed k-means centroids sampling strategy (on the right), starting from the same $N = 20$ initial \textit{guesses}.} \label{fig:sampling}
\end{figure}
The sampling approach described above represents the common way to draw new samples from the learned prior of a VQ-VAE. However, we build upon it to create a stronger and richer selection strategy that furthers the multi-modal capabilities of DDPMs. The standard evaluation process involves sampling $K$ distinct trajectories from the model and assessing the top-performing one (as described in Sec.~\ref{sec:experiments}). Therefore, each methodology must find the right balance between accuracy in its prediction and potential for exploration. The proposed procedure goes in this direction: we generate numerous \textit{raw} future paths, called \textit{guesses}, and then condense them into the most representative ones. In formal terms, we sample $N$ guesses and then perform the k-means clustering algorithm, with a number of clusters equal to $K < N$ (in our experiments, we set $N=200$ and $K=20$). We view the resulting \textit{centroids} as the principal modes of the predictive distribution learned by the DDPM and thus use them for prediction in place of the original samples. This strategy guarantees a twofold advantage compared to naive prediction: firstly, out-of-distribution samples typically form independent clusters, thus enhancing exploration; secondly, the use of centroids reduces the quantization noise, as in-distribution samples are grouped into large clusters and averaged element-wise (see Fig.~\ref{fig:sampling}).
\section{Experiments}
\label{sec:experiments}
We assess our proposal on the following three trajectory prediction benchmarks.
\tit{Stanford Drone Dataset (SDD).}The dataset~\cite{robicquet2016learning} gathers trajectories of pedestrians within the Stanford University campus in a bird’s eye view. Given $8$ time steps ($\approx 3.2$ seconds), methods have to forecast the subsequent $12$ frames ($4.8$ seconds). We employ the established train-test split~\cite{mangalam2020not}.

\tit{NBA SportVU Dataset (NBA).}Collected by the NBA's SportVU automatic tracking system, this dataset~\cite{NBA} provides the trajectories of $10$ players and the ball in real basketball games. Given $10$ previous time-steps ($\approx 2.0$ seconds), the models predict the subsequent $20$ steps ($4.0$ seconds).

\tit{NFL Football Dataset (NFL).}The NFL Football Dataset~\cite{NFL} records the movements of every player throughout each play of the $2017$ season. The goal is to predict the trajectories of the $22$ players ($11$ per team) and the ball for the ensuing $3.2$ seconds ($16$ steps), given the preceding $1.6$ seconds ($8$ steps).

\smallskip
\tit{Metrics.}We use two established metrics~\cite{pellegrini2009you,alahi2016social} \ie the Average/Final Displacement Errors (ADE/FDE). Given predicted and ground-truth trajectories, ADE computes the average error on all points, while the FDE restricts the error committed in the final step. Following other works dealing with stochastic models~\cite{ivanovic2019trajectron,mangalam2020not}, we adhere to the best-of-$20$ protocol~\cite{xu2022remember,xu2022groupnet}, selecting for evaluation the best trajectory from a pool of $K=20$ generations. We denote the corresponding metrics as ADE$_{K}$ and FDE$_{K}$; these are in meters for NBA and NFL, and in pixels for SDD. For sports datasets, we compute these metrics at different delta times to provide a more comprehensive assessment.

\smallskip
\tit{Implementation Details\footnote{The code is available at~\url{https://github.com/aimagelab/LRVQ}.}.}We set the number of codewords $C$ to $16$ for all datasets, while we take the best rank $r$ for each dataset (\eg 8 for SDD and NBA, 4 for NFL). For the first stage, we use AdamW~\cite{loshchilov2017decoupled} as optimizer with $\operatorname{lr} = 5 \times 10^{-4}$, $\beta_1=0.5$ and $\beta_2=0.9$. We train on SDD for $7000$ epochs with batch size equal to $256$. For NBA and NFL, we instead optimize for $700$ epochs (the batch size equals $64$). We use a cosine schedule for $\lambda_\xi$ from an initial value of $0$ to a final value of $1$. In this way, we can introduce the instance-level codebook gradually during training.

For the second stage, we re-use the same optimizer/batch-size setup, while training for $3000$ epochs for SDD, $1000$ epochs for NBA, and $700$ for NFL. As an augmentation technique, we rotate the trajectories by a random angle, ranging between $0$ and $\theta_{max}$. We set $\theta_{max}$ to $180^\circ$ for the first stage, while we find it beneficial to adopt a lower value ($5^\circ$) for the second stage. 
\begin{table*}[t]
\centering
\setlength{\tabcolsep}{0.8em}{\caption{Impact of distinct VQ-VAE codebooks on performance (ADE$_{20}$/FDE$_{20}$).}
\renewcommand{\arraystretch}{1.08}
\begin{tabular}{c|ccc}
\hline
\rule{0pt}{\normalbaselineskip}
    \hspace{-0.4em}Dataset
    & Static
    & Full-Rank
    & Low-Rank
    \\
\hline
\rule{0pt}{\normalbaselineskip}
     \hspace{-0.4em}SDD 
     & $8.29/13.44$
     & $8.07/12.89$
     & $\boldsymbol{7.86}/\boldsymbol{12.68}$\\
\rowcolor{lightgray}
\rule{0pt}
{\normalbaselineskip}
     \hspace{-0.4em}NBA 
     & $0.895/1.279$ 
     & $0.894/1.275$ 
     & $\boldsymbol{0.893}/\boldsymbol{1.267}$\\
     \rule{0pt}{\normalbaselineskip}
     \hspace{-0.4em}NFL 
     & $0.993/1.702$
     & $0.993/1.702$
     & $\boldsymbol{0.982}/\boldsymbol{1.679}$\\
\hline
\end{tabular}
\label{table:st_dyn_lr}}
\end{table*}
\subsection{On the Impact of the Instance-based Codebook}
\label{sec:effective codebook}
To assess the merits of our \textit{low-rank} instance-based codebook, we herein empirically compare it with two alternative strategies. On the one hand, we devise a comparison with a \textit{static} codebook ($\rightarrow$ standard VQ-VAEs, lacking instance-level conditioning). Secondly, we contrast it with a \textit{full-rank} codebook (which includes instance-level conditioning but lacks low-rank design constraints). To be more precise, the \textit{full-rank} codebook is a baseline approach herein provided, which computes the values of the codebook through a learnable module fed with historical information as input. Unlike the proposed \textit{low-rank} counterpart, the \textit{full-rank} codebook does not adapt a shared static codebook but directly outputs its values. Through such a comparison, we can evaluate the efficacy of constraining the updates to the dictionary within a low-dimensional manifold.

Tab.~\ref{table:st_dyn_lr} presents the related results: as can be observed, the \textit{low-rank} model outperforms both the \textit{static} and \textit{full-rank} variants. In particular, the improvements are remarkable for SDD and NFL and more modest for NBA. Moreover, the presence of instance-level conditioning, common to \textit{full-} and \textit{low-} approaches, proves particularly beneficial for the SDD dataset, as demonstrated by the gap w.r.t. the static codebook (similar evidence emerges for the NBA dataset).
\begin{table*}[t]
\centering
\setlength{\tabcolsep}{0.8em}{\caption{Impact of varying the rank of $B$ on the behavior of the model. 
Optimal performance (ADE$_{20}$) is achieved by identifying a sweet spot characterized by a low reconstruction error ($\mathrm{ADE_{rec}}$) and a high accuracy in code prediction (Acc).}
\renewcommand{\arraystretch}{1.08}
\begin{tabular}{c|c|ccc@{\hspace{0.8em}}c@{\hspace{0.8em}}}
\hline
\rule{0pt}{\normalbaselineskip}
    \hspace{-0.4em}Dataset
    & Rank
    & $\mathrm{ADE_{rec}}$ $\downarrow$
    & Acc(\%) $\uparrow$ & ADE$_{20}$ $\downarrow$
    \\
\hline
\rule{0pt}{\normalbaselineskip}
    \multirow{2}{*}{\hspace{-0.4em}SDD}
     & 4 & 3.41 & 26.38 & 7.96
     \\
     & \cellcolor{lightgray}16 & \cellcolor{lightgray}2.97 & \cellcolor{lightgray}22.20 & \cellcolor{lightgray}8.06
     \\
     \hline
\rule{0pt}{\normalbaselineskip}
     \multirow{2}{*}{\hspace{-0.4em}NBA}
     & 4 & 0.207 & 15.92 & 0.898 
     \\
     & \cellcolor{lightgray}16 & \cellcolor{lightgray}0.164  & \cellcolor{lightgray}13.27 & \cellcolor{lightgray}0.892
     \\
     \hline
     \rule{0pt}{\normalbaselineskip}
     \multirow{2}{*}{\hspace{-0.4em}NFL}
     & 4 & 0.227 & 15.30 & 0.982
     \\
     & \cellcolor{lightgray}16 & \cellcolor{lightgray}0.177 & \cellcolor{lightgray}11.95 & \cellcolor{lightgray}0.996
     \\
\hline
\end{tabular}
\label{table:var_r}}
\end{table*}

In the second place, we aim to investigate the impact of the rank $r$, which controls the dimension of the matrix $B_{\mathrm{ctx}}$ (\ie the degree of instance-level cues introduced into the codebook). In particular, we want to measure how the rank $r$ affects: \textit{i)} the reconstruction capabilities of the VQ-VAE decoder (learned during the first stage); \textit{ii)} the generative capabilities of the diffusion model (learned during the second stage). For point \textit{i)}, we exploit the Average Displacement Error ($\mathrm{ADE_{rec}}$) to assess the reconstruction performance. Instead, to characterize the generative capabilities, we resort to the mean accuracy achieved by the diffusion model in predicting codebook indexes, as well as the already mentioned ADE$_{20}$.

Tab.~\ref{table:var_r} presents the results for different ranks $r$. We observe that a higher reconstruction capability during the initial training stage is associated with increased difficulty in the diffusion task, resulting in lower accuracy. This indicates a correlation between the two phases: achieving optimal results in the first phase does not necessarily yield the best final generation metrics, as it complicates the joint task of trajectory generation (\ie sampling from the prior and reconstructing through the decoder). Tab.~\ref{table:var_r} demonstrates that the most favorable final metrics are achieved by striking a balance between low reconstruction error and good diffusion accuracy.
%
\subsection{Comparison with SOTA Methods}
In this section, we compare our model to the following existing approaches:
\begin{itemize}
    \item Social-GAN~\cite{gupta2018social} relies on a Conditioned GAN, with a module to handle social interactions between agents.
    \item Trajectron\texttt{++}~\cite{salzmann2020trajectron++} exploits VAEs and graph-structured recurrent networks.
    \item PECNet~\cite{mangalam2020not} augments a VAE with \textit{goal-oriented} reasoning.
    \item LB-EBM~\cite{pang2021trajectory} targets the prediction of long-range trajectories through a belief vector, which encapsulates the energy distribution in the environment.
    \item GroupNet~\cite{xu2022groupnet} is a multiscale hypergraph network that captures both pair- and group-wise interactions at different scales. 
    \item Memo-Net~\cite{xu2022remember} mimics retrospective memory in neuropsychology and predicts intentions by retrieving similar instances from a memory bank.
    \item MID~\cite{gu2022stochastic} leverages a diffusion model to progressively reduce indeterminacy within potential future paths.
\end{itemize}
We report the comparison in Tab.~\ref{table:sdd}, Tab.~\ref{table:nba}, and Tab.~\ref{table:nfl}. To sum up, our \methnam demonstrates superior performance across all the considered benchmarks.

On the SDD dataset (Tab.~\ref{table:sdd}), we attain superior ADE results, matching closely MemoNet in FDE. While PECNet and GroupNet, among C-VAE methods, demonstrate noteworthy performance compared to the older S-GAN and Trajectron\texttt{++}, they struggle in FDE, especially when compared to MemoNet. This could be ascribed to the effective sampling strategy of MemoNet, which integrates a tailored clustering phase to generate multiple overall intentions.

Additionally, our approach showcases robust performance across all examined partial timestamps for both the NBA (Tab.~\ref{table:nba}) and NFL datasets (Tab.~\ref{table:nfl}). The two most competing methods are GroupNet -- based on the C-VAE framework -- and more importantly MID, which akin to our approach utilizes a diffusion process. However, we highlight an important distinction with MID, which we consider as a motivation for our improvements: while MID adopts diffusion modeling directly in output space, we instead apply it to the discrete variables extracted by the VQ-VAE encoder. We believe that our latent-based formulation further promotes the emergence of multi-modal generative capabilities.
\begin{table*}[!t]
\footnotesize
\centering
\setlength{\tabcolsep}{1mm}{\caption{\small SDD results (ADE$_{20}$/FDE$_{20}$). $^\ast$ represents the reproduced results from open source. Best results in \textbf{bold}, second-best \underline{underlined}.}
\fontsize{7.8}{9.4}\selectfont
\renewcommand{\arraystretch}{1.08}
\setlength{\tabcolsep}{0.3em}
\begin{tabular}{c|cc@{\hspace{0.6em}}c@{\hspace{0.6em}}ccccccc|c}
\hline
\rule{0pt}{\normalbaselineskip}
    Time
    &S-GAN
    &Trajectron\tiny{++} 
    &PECNet
    &MemoNet
    &GroupNet
    &MID$^\ast$
    &\textbf{\methnam}
    \\
    \hline
    \rule{0pt}{\normalbaselineskip}
     \hspace{-0.4em}4.8$s$ 
     & $27.23/41.44$
     & $19.30/32.70$ 
     & $9.96/15.88$ 
     & $\underline{8.56}/\boldsymbol{12.66}$
     & $9.31/16.11$ 
     & $9.73/15.32$ 
     & $\boldsymbol{7.86}/\underline{12.68}$\\
\hline
\end{tabular}

\label{table:sdd}}
\end{table*}
\begin{table*}[!t]
\footnotesize
\centering
\setlength{\tabcolsep}{1mm}{\caption{\small NBA results (ADE$_{20}$/FDE$_{20}$). Best results in \textbf{bold}, second-best \underline{underlined}.}
\fontsize{7.8}{9.8}\selectfont
\renewcommand{\arraystretch}{1.08}
\setlength{\tabcolsep}{0.5em}{
\rowcolors{2}{lightgray}{}
\begin{tabular}{c|cc@{\hspace{0.6em}}c@{\hspace{0.6em}}ccccccc|c}
\hline
\rule{0pt}{\normalbaselineskip}
    Time
    &S-GAN
    & PECNet
    &Trajectron\tiny{++}
    &MemoNet
    &GroupNet
    &MID
    &\textbf{\methnam}
    \\
    \hline
    \rule{0pt}{\normalbaselineskip}
    \hspace{-0.4em}$1.0s$
     & $0.41/0.62$ 
     & $0.40/0.71$ 
     & $0.30/0.38$ 
     & $0.38/0.56$ 
     & $\underline{0.26}/\underline{0.34}$ 
     & $0.28/0.37$ 
     & $\boldsymbol{0.19}/\boldsymbol{0.29}$
     \\
     $2.0s$ 
     & $0.81/1.32$ 
     & $0.83/1.61$ 
     & $0.59/0.82$ 
     & $0.71/1.14$ 
     & $\underline{0.49}/\underline{0.70}$
     & $0.51/0.72$ 
     & $\boldsymbol{0.41}/\boldsymbol{0.63}$
     \\
     $3.0s$ 
     & $1.19/1.94$ 
     & $1.27/2.44$ 
     & $0.85/1.24$ 
     & $1.00/1.57$ 
     & $0.73/1.02$ 
     & $\underline{0.71}/\underline{0.98}$ 
     & $\boldsymbol{0.64}/\boldsymbol{0.96}$
     \\
     $4.0s$ 
     & $1.59/2.41$ 
     & $1.69/2.95$ 
     & $1.15/1.57$ 
     & $1.25/1.47$ 
     & $\underline{0.96}/\underline{1.30}$ 
     & $\underline{0.96}/\boldsymbol{1.27}$
     & $\boldsymbol{0.89}/\boldsymbol{1.27}$
     \\
\hline
\end{tabular}
}

\label{table:nba}}
\end{table*}
\begin{table*}[!t]
\footnotesize
\centering
\setlength{\tabcolsep}{1mm}{\caption{\small 
NFL results (ADE$_{20}$/FDE$_{20}$). Best results in \textbf{bold}, second-best \underline{underlined}.}
\fontsize{7.8}{9.8}\selectfont
\renewcommand{\arraystretch}{1.08}
\setlength{\tabcolsep}{0.58em}
\rowcolors{2}{lightgray}{}
\begin{tabular}{c|cc@{\hspace{0.6em}}c@{\hspace{0.6em}}ccccccc}
\hline
    \rule{0pt}{\normalbaselineskip}
    \hspace{-0.4em}Time
    &S-GAN
    & PECNet
    & Trajectron\tiny{++}
    & LB-EBM
    & GroupNet
    & MID
    &\textbf{\methnam}
    \\
    \hline
    \rule{0pt}{\normalbaselineskip}
\hspace{-0.4em}$1.0s$   & $0.37/0.68$ 
                & $0.52/0.97$ 
                & $0.41/0.65$ 
                & $0.75/1.05$ 
                & $0.32/\underline{0.57}$ 
                & $\underline{0.30}/0.58$ 
                & $\boldsymbol{0.23}/\boldsymbol{0.35}$ \\
$2.0s$            & $0.83/1.53$ 
                & $1.19/2.47$ 
                & $0.93/1.65$ 
                & $1.26/2.28$ 
                & $0.73/1.39$ 
                & $\underline{0.71}/\underline{1.31}$ 
                & $\boldsymbol{0.53}/\boldsymbol{0.92}$ \\
$3.2s$     & $1.44/2.51$ 
                & $1.99/3.84$ 
                & $1.54/2.58$ 
                & $1.90/3.25$ 
                & $1.21/2.15$ 
                & $\underline{1.14}/\underline{1.92}$ 
                & $\boldsymbol{0.98}/\boldsymbol{1.68}$ \\
\hline
\end{tabular}
\label{table:nfl}}
\end{table*}
\begin{figure}[t]
\centering
    \includegraphics[width=0.95\textwidth]{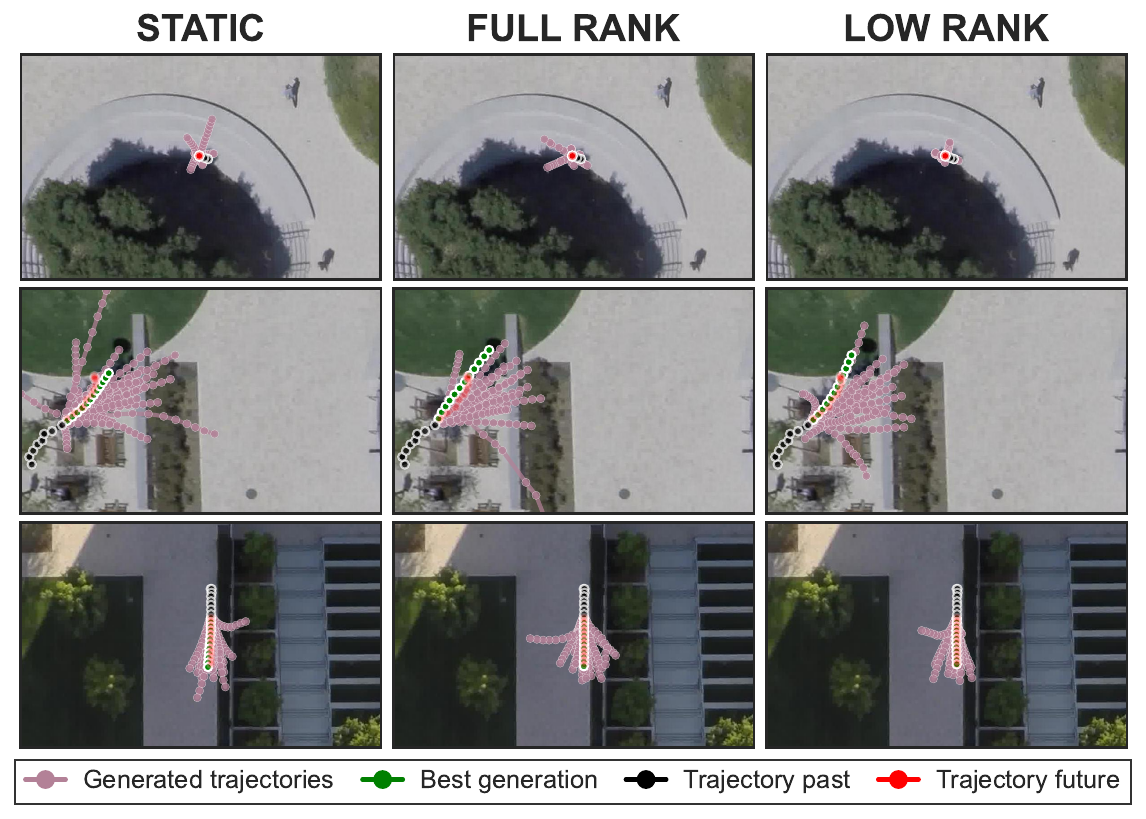}
    \caption{Qualitative comparison for three SDD scenes (one for each row of the figure) between the trajectories obtained from a VQ-VAE with a static codebook, a full rank codebook the proposed \textit{low-rank} codebook (from left to right).} \label{fig:qualitative}
\end{figure}
\subsection{Qualitative Results}
Figure~\ref{fig:qualitative} provides a qualitative comparison on $20$ generations (with sub-sampling) produced by a VQ-VAE trained with a \textit{static} codebook, a \textit{dynamic} codebook, and the \textit{low-rank} conditioned codebook (see Sec.~\ref{sec:effective codebook}). Each row illustrates a different scene from the SDD dataset, showcasing different agent behaviors: in the first one, the agent remains stationary, while in the others, it either turns left or proceeds straight ahead. Compared to the other two methods, low-rank conditioning appears to be more accurate, particularly in complex scenarios where the agent stays still or changes its direction of movement.
\section{Discussion and Conclusions}
\tit{Limitations.} The complexity of our model is linked to two factors:
\begin{itemize}
\item Two-step training procedure: although VQ-VAE offers benefits such as a learned prior, the training must be divided into two distinct stages, which increases the total time required to train the model.
\item Inference time: the inference procedure described in Sec.~\ref{sec:multimodal} takes longer as the number $N$ of starting guesses increases. To obtain a trade-off between the accuracy of the ensemble of $K$ final generations and the computational time, the parameter $N$ has to be carefully adjusted.
\end{itemize}

\tit{Conclusion.} We propose a stochastic approach for trajectory prediction. It builds upon Vector Quantization to yield a predictive distribution that preserves both sampling fidelity and diversity. Our main contribution lies in a dynamic, instance-related codebook encompassing past trajectory information. Notably, contextual information is incorporated into the codebook through a low-rank update. We conduct several empirical studies to validate our approach, demonstrating its superior generative capabilities compared to both standard VQ-VAEs and existing methods. This leads to state-of-the-art results on three established benchmarks.
\section*{Acknowledgement}
The research was supported by the Italian Ministry for University and Research through the PNRR project ECOSISTER ECS 00000033 CUP E93C22001100001 and by the EU Horizon project “ELIAS - European Lighthouse of AI for Sustainability” (No. 101120237).
%
%
\bibliographystyle{splncs04}
\bibliography{full,egbib}
\end{document}